\documentclass[11pt,a4paper]{article}
\usepackage[hyperref]{eacl2021}
\usepackage{times}
\usepackage{latexsym}

\usepackage{graphicx}
\usepackage{amssymb}
\usepackage{amsmath}
\usepackage{multirow}
\usepackage{booktabs}
\usepackage{todonotes}
\usepackage{microtype}

\aclfinalcopy % Uncomment this line for the final submission

\title{Effective Distant Supervision for Temporal Relation Extraction}

\author{Xinyu Zhao\qquad Shih-ting Lin\qquad Greg Durrett \\
        Department of Computer Science \\
        The University of Texas at Austin\\
        \texttt{\{xinyuzhao,j0717lin\}@utexas.edu\qquad gdurrett@cs.utexas.edu}}

\date{}

\begin{document}
\maketitle
\begin{abstract}
A principal barrier to training temporal relation extraction models in new domains is the lack of varied, high quality examples and the challenge of collecting more. We present a method of automatically collecting distantly-supervised examples of temporal relations. We scrape and automatically label event pairs where the temporal relations are made explicit in text, then mask out those explicit cues, forcing a model trained on this data to learn other signals. We demonstrate that a pre-trained Transformer model is able to transfer from the automatically labeled examples to human-annotated benchmarks in both zero-shot and few-shot settings, and that the masking scheme is important in improving generalization.\footnote{Code and datasets available at: \url{https://github.com/xyz-zy/distant-temprel}}
\end{abstract}

\section{Introduction}

Temporal relation extraction has largely focused on identifying pairwise relationships between events in text. Past work on annotating temporal relations has struggled to devise annotations schemes which are both comprehensive and easy to judge \cite{putejovsky-2003-timebank,cassidy-etal-2014-annotation}. However, even simplified annotation schemes designed for crowdsourcing \cite{Ning:2018,vashishtha-etal-2019-fine} can struggle to acquire high-accuracy judgments about nebulous phenomena, leading to a scarcity of high-quality labeled data. Compared to tasks like syntactic parsing \cite{bies-etal-2012-english} or natural language inference \cite{williams-etal-2018-broad}, there are thus fewer resources for temporal relation extraction in other domains.

In this work, we present a method of automatically gathering distantly-labeled temporal relation examples. Unlike traditional distant supervision methods \cite{Mintz:09}, we do not rely on a knowledge base, but instead on heuristic cues that we will then mask out, forcing the model to make inferences from the remaining context. We explore two types of cues, but focus primarily on events that are anchored to orderable timexes \citep{goyal-durrett-2019-embedding}. These examples can be collected and labeled using an automatic system \cite{Chambers:2014}. By then masking the explicit temporal indicators, a model trained on these examples can no longer learn trivial timex-based rules, but must instead attend to more general temporal context cues. We show that a pre-trained model fine-tuned on this data learns general, implicit cues that transfer more broadly to human-annotated benchmarks. This observation follows a trend of recent work showing pre-trained models' ability to generalize from synthetic data to natural data \citep{xu-etal-2020-autoqa, marzoev2020unnatural}.

We implement our approach with pre-trained Transformer models \citep{devlin-etal-2019-bert, liu-etal-2019-robustly, clark-etal-2020-electra} similar to a state-of-the-art temporal relation extraction model from the literature \citep{han-etal-2019-joint}. Our model is able to effectively transfer from a distantly-labeled dataset to the MATRES benchmark \cite{Ning:2018} when used to supplement a small number of in-domain or out-of-domain samples.

\section{Classification Model}
\label{sec:model}

\begin{figure}[]
    \centering
    \includegraphics[width=7.5cm]{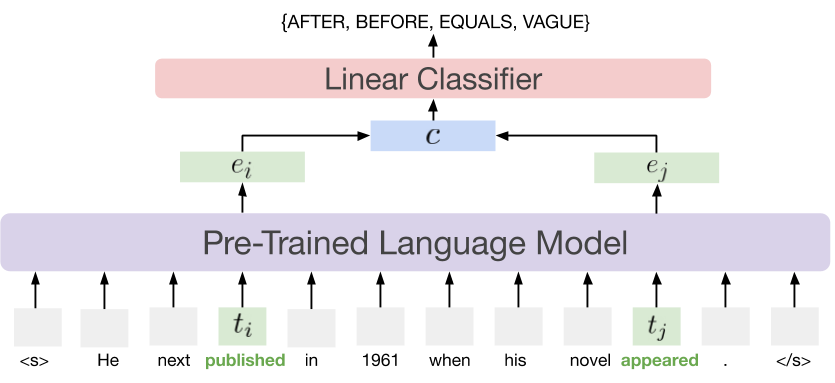}
    \caption{Classification model consisting of a pre-trained transformer model and a linear layer. Event tokens are represented by $t_i, t_j$ and correspond to output embeddings $e_i, e_j$, which are used by the linear classifier to produce a distribution over labels.}
    \label{fig:cls-model}
\end{figure}

Our base classification model consists of a pre-trained Transformer \cite{vaswani} model with an appended linear classification layer, represented in Figure \ref{fig:cls-model}. For the majority of our experiments, we use RoBERTa \citep{liu-etal-2019-robustly}, which we found to work better than BERT \cite{devlin-etal-2019-bert} and ELECTRA \citep{clark-etal-2020-electra} for domain transfer. We chose a single set of hyperparameters by tuning to match the performance of \citet{han-etal-2019-joint}; for details see Appendix \ref{sec:impl}.

%%%%%%%%%%%%%%%
%%%%%%%%%%%%%%%

A single example consists of an event pair in context, which may be a single sentence or two sentences, and a label from \{\textsc{after, before, equals, vague}\}, following the annotation scheme in MATRES.
Each example is tokenized to yield input tokens $T = [t_1, t_2, ..., t_n]$, with event tokens $t_i, t_j \in T$. For events consisting of multiple sub-word tokens, we track only the first token position. We use the convention of passing events in text order, where $1 \leq i < j \leq n$.

The language model then produces output embeddings $[e_1, e_2, ..., e_n]$. For classification, we select the embeddings $e_i, e_j$ corresponding to the event token positions, and combine them into a classification vector, $c = [e_i ; e_j; e_i \odot e_j ; e_i - e_j]$ where $\odot$ represents elementwise multiplication. Finally, a linear classification layer produces a distribution over the four relation labels. Training is done by maximizing likelihood of labeled samples. We implement this model using PyTorch \citep{pytorch} and pre-trained models from HuggingFace's Transformers library \citep{transformers}.

We benchmark our classification model by training and evaluating on MATRES. We achieve an F1 of 79.8 with RoBERTa \cite{liu-etal-2019-robustly} and an F1 of 80.3 with ELECTRA \cite{clark-etal-2020-electra}, demonstrating that our model replicates state-of-the-art performance achieved by local models (only considering arcs in isolation), currently 80.3 F1 \citep{han-etal-2019-joint}.

\section{Learning from Distant Data}
\label{sec:learning-distant-data}

We aim to create a method of automatically gathering high-quality data that can be applied to unlabeled text. To this end, we focus on two techniques identifying explicit temporal indicators. First, we identify single-sentence examples where event pairs are automatically labeled via explicit discourse connectives. Second, we scrape occurrences of event pairs that are anchorable to timexes which determine their relation. We will see that this second technique is substantially better, and analyze some factors contributing to the performance delta. Although neither technique captures the gamut of phenomena found in human-labeled data, pre-trained models' generalization capabilities and a masking technique tailored for this setting are two tools that enable effective transfer.

For both techniques, we scrape distant examples from English Gigaword Fifth Edition \citep{parker-etal-2011-english}. We extract samples from a balance of the different news sources present in the dataset. In both cases, we use the Stanford CoreNLP lexicalized parser \cite{manning-EtAl:2014:P14-5} to generate parse trees for the source text, which can be time-consuming at scale. However, we can pre-filter sentences based on the presence of timexes or target discourse connectives, and so in practice we only rarely need to invoke the parser.  Table~\ref{distant-exs} in the Appendix shows collected data samples, and we describe these two collection methods in more detail below.

\subsection{Temporal Connectives}

Words like \emph{before}, \emph{after}, \emph{during}, \emph{until}, \emph{prior to}, and others can indicate the temporal status of events in text explicitly. Past work has shown that complex relations can be learned from discourse connectives in non-temporal settings  \citep{nie-etal-2019-dissent}, so such connectives can be powerful indicators. We focus on \emph{before} and \emph{after} in this work, as these are the most common and straightforward to map to a temporal relation.

To identify connected event pairs, we use the Stanford CoreNLP lexicalized parser \cite{manning-EtAl:2014:P14-5} to produce parse trees. We then search for a related event pair by 1) identifying the connective, 2) finding the closest parent verb phrase, and 3) finding the closest child verb phrase. These become the events for the example. When this identifies modals or auxiliaries, we take the corresponding main verb. The label for the example is simply determined by the \emph{before} or \emph{after} connective. Examples are listed in Appendix \ref{sec:distant-examples}; on inspection, we found this method to be reliable.

\subsection{Events Anchored to Time Expressions}
\label{sec:distant-timex}

\begin{figure}[]
    \centering
    \includegraphics[width=7.5cm]{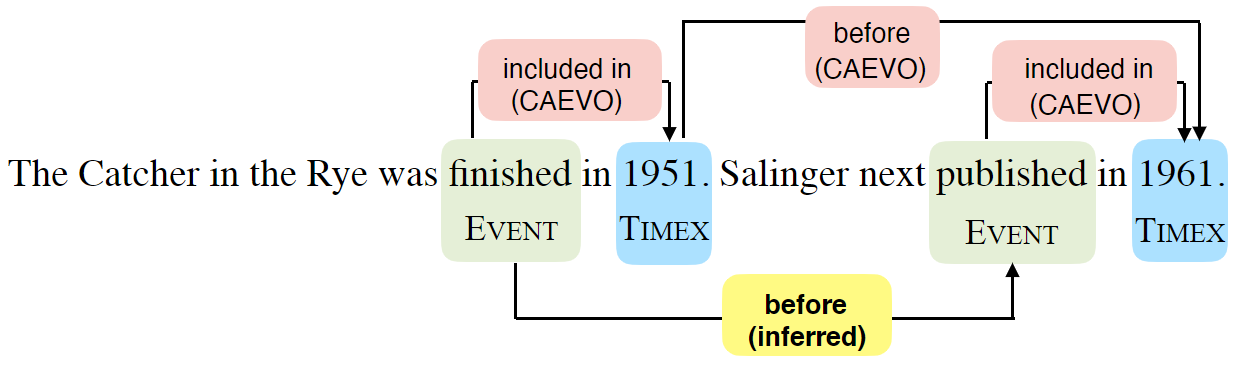}
    \caption{Process of identifying distantly labeled event-event relations using CAEVO. First, events and timexes are identified and two of CAEVO's sieves are applied. Then, using transitivity, event-event relation is inferred from associated timexes.}
    \label{fig:caevo_parse}
\end{figure}

Beyond connectives, another cue is the explicit presence of timexes. An example is shown in Figure~\ref{fig:caevo_parse}: the years \emph{1951} and \emph{1961} determine the ordering of their associated events, assuming each event can be appropriately linked to the timex.

We use CAEVO \citep{Chambers:2014} to detect event pairs and link events to timexes, which include explicit datetimes (\textit{January}, \textit{1961}), relative times (\textit{tomorrow}) and other natural language indicators (\textit{now}, \textit{until recently}). This approach, which yields both single- and cross-sentence examples, was explored by \citet{goyal-durrett-2019-embedding}, who noisily labeled data to evaluate their timex embedding model.

First, the input document is annotated by CAEVO with events and timexes using its parse trees. Two of its sieves are then applied: the \textsc{AdjacentVerbTimex} sieve identifies events that are anchored to time expressions via a direct path in the syntactic parse tree,  then the \textsc{TimeTimeSieve} uses a small set of deterministic rules to label relations between timexes. These two sieves have high precisions, of 0.74 and 0.90, respectively \cite{Chambers:2014}. Figure~\ref{fig:caevo_parse} shows the result of applying both sieves. Finally, the system is able to infer the relations between events that are anchored to comparable timexes (i.e \textit{finished} before \textit{published}), giving us event pairs usable for training.

The resulting datasets are reasonably balanced between \textsc{before} and \textsc{after}, with sparse \textsc{equal} examples and \emph{no} \textsc{vague} examples. A more detailed label breakdown is included in Appendix \ref{sec:label-breakdown}.

%%%%%%%%%%%%%%%
%%%%%%%%%%%%%%%

\subsection{Example Masking}

These distant examples are gleaned from ``trivial'' indicators in the text, which a model like BERT \citep{devlin-etal-2019-bert} will overfit to. We observe that our RoBERTa classifier yields 99.8\% accuracy on a held out BeforeAfter dataset, and evaluating with the same DistantTimex train/test split of \citet{goyal-durrett-2019-embedding} results in 96.6\% test accuracy.

In order to combat this, we \emph{mask} the explicit temporal cues identified by our weak labeling process. Our goal is to induce the model to learn the label from the remaining tokens, including the event instances themselves and the broader context. Masking is performed \emph{prior to} subword tokenization, so each word or timex gets one mask token per word. For our BeforeAfter examples, we simply mask the temporal connective. For our DistantTimex examples, we use the timex tags generated by CAEVO to mask all identified timexes present in the context. This results in masking of not only explicit timexes (e.g. dates, times), but also of natural language timexes (e.g. \textit{previously}, \textit{recently}). This may occasionally result in ``sparse" training examples that have a high proportion of mask tokens. Refer to Tables 7-8 in Appendix \ref{sec:masking} for concrete examples.

In spite of this masking, our model is able to classify distantly-labeled timex examples with 85\% accuracy when evaluating on a held-out set, well above a majority baseline. This indicates that there are other temporal cues that the model can use to determine the temporal relation.

\section{Results}

We evaluate our distant dataset on several axes. (1) In zero-shot, few-shot, or transfer settings, to what extent can this help existing models? (2) How important is each facet of our distant extraction setup? (3) What can we say about the distribution of our data from these techniques? We focus our evaluation on the English MATRES dataset \cite{Ning:2018}, a four-class temporal relation dataset of chiefly newswire data drawn from a number of different sources.\footnote{We choose \textbf{not} to evaluate on the UDS-T dataset \cite{vashishtha-etal-2019-fine}, treating it solely as a training source. In our experiments, converting from real-valued time span annotations into categorical event-pair labels required dealing with significant disagreement among annotators. Despite trying several resolution strategies, none of our \emph{in-domain} fine-tuned Transformer models performed much better than a majority baseline, indicating high noise or an extremely challenging setting.}

We found significant variance in model performance in our transfer for small data settings, so most results use average best performance or majority-vote ensembled performance of three randomly seeded models trained in the same setting.

\paragraph{Comparison of Distant Datasets}
To get the clearest picture of the differences in our distant setups, we first evaluate in a setting of zero-shot adaptation to MATRES. We train on different distantly labeled dataset sizes in order to establish a relationship between example quantity and generalization performance. Our results presented in Table \ref{distant-scale} show that the DistantTimex data works substantially better than BeforeAfter: with the DistantTimex data, there is a correlation between adding more distantly-labeled examples and increased performance. While both of these rules target particular narrow slices of data, the set with explicit timexes appears to be broader than that with before/after connectives, and hence BERT can learn to generalize better.\footnote{One possible reason is that explicit indicators like \emph{before} and \emph{after} may be used explicitly to communicate temporal information where it cannot be otherwise inferred, but timexes are often used to communicate more specific details about events where the relation may already be clear.}

%%%%%%%%%%%%%%%
\begin{table}[]
\small
\centering
\begin{tabular}{ll l l l l}
\toprule
  & \multirow{2}{*}{} &
\multicolumn{4}{c}{\textbf{MATRES}} \\ 
\multicolumn{2}{l}{\textbf{Training Data}} & \textbf{Split} & \textbf{P} & \textbf{R} & \textbf{F1} \\ \midrule

\multirow{2}{*}{Majority Label} & & Dev  & 52.6 & 60.2 & 56.1 \\
                              &      & Test & 50.7 & 58.6 & 54.3  \\
                              \midrule
\multirow{2}{*}{MATRES}       &     & Dev  & 77.1 & 85.5 & 81.1 \\ 
                              &     & Test & 75.1 & 84.8 & 79.6 \\ \midrule
\multirow{6}{*}{DistantTimex} & 1k  & Dev  & 54.1 & 61.9 & 57.7 \\
                              & 1k  & Test & 50.8 & 58.7 & 54.5 \\
                              & 5k  & Dev  & 60.4 & 69.1 & 64.5 \\  
                              & 5k  & Test & 61.9 & 71.5 & 66.4 \\  
                              & 10k & Dev  & 64.0 & 73.2 & 68.3 \\
                              & 10k & Test & 61.5 & 71.1 & 66.0 \\ \midrule
\multirow{2}{*}{BeforeAfter}  & 10k & Dev  & 53.4 & 61.1 & 57.0 \\ %Epoch 3
                              & 10k & Test & 51.6 & 59.7 & 55.3 \\  \bottomrule
\end{tabular}
\caption{Results on MATRES after training on distant data, with explicit temporal cues masked. Results are esembled from three models trained in the same setting.}
\label{distant-scale}
\end{table}
%%%%%%%%%%%%%%%

\paragraph{Using Fewer Labeled Examples}

%%%%%%%%%%%%%%%
\begin{table}[]
\small
\centering
\setlength{\tabcolsep}{4pt} % Default value: 6pt
\begin{tabular}{ll l l l l}
\toprule
  & \multirow{2}{*}{} &
            \multicolumn{4}{c}{\textbf{DistantTimex Examples}} \\ 
\textbf{Labeled Set} & \textbf{Eval}               & \textbf{None}           & \textbf{1k}           & \textbf{5k}     &\textbf{10k}    \\ \midrule
                            
\multirow{4}{*}{MATRES 1k}  & Avg. Dev F1  & 64.2  & 67.4  & 75.2  & 75.6  \\ 
                            & Avg. Test F1 & 60.9  & 66.1  & 73.6  & 73.7  \\
                            
  & Ens. Dev F1  & 70.2  & 74.5  & 76.7  & 76.6  \\ 
                            & Ens. Test F1 & 66.5  & 72.0  & 75.0  & 75.5  \\ 
                            \midrule
\multirow{4}{*}{UDS-T 5k} & Avg. Dev F1   & 68.7 & 62.8 & 72.0  & 70.8 \\
                          & Avg. Test F1  & 66.2 & 60.9 & 69.5  & 69.8 \\
                          & Ens. Dev F1   & 70.1 & 64.6 & 73.2  & 72.1 \\
                          & Ens. Test F1  & 68.2 & 62.3 & 70.7  & 71.8 \\
                          \bottomrule

\end{tabular}
\caption{Evaluation results on MATRES when adding automatically collected examples to small amounts of human-annotated training data. Using more DistantTimex data is able to improve performance substantially over not using any (\textbf{None}).}
\label{matres-plus-distant}
\vspace{-2mm}
\end{table}
%%%%%%%%%%%%%%%

In a more realistic setting, we assume access to small amounts of pre-existing labeled data, using roughly 10\% of existing datasets. Table \ref{matres-plus-distant} shows results from evaluating on MATRES using small amounts of either MATRES or UDS-T data in conjunction with our distant data; three models are randomly initialized and trained for each setting. In both settings, adding distant data improves substantially over just using the in-domain MATRES data, and the best model performance is only around 4 F$_1$ worse than the in-domain MATRES results using the entire train set. We also show that this data can stack with data from UDS-T \cite{vashishtha-etal-2019-fine} and improve transfer over raw UDS-T. This is in spite of very different event distributions between UDST and MATRES and a complete lack of examples of \textsc{vague} relations during training.

%%%%%%%%%%%%%%%
\begin{table}[]
\small
\centering
\begin{tabular}{ll l l l}
\toprule
  & \multirow{2}{*}{} & \multicolumn{3}{c}{\textbf{MATRES Test}} \\ 
\multicolumn{2}{l}{\textbf{Training Data}} & \textbf{P} & \textbf{R} & \textbf{F1} \\ \midrule

%MATRES                &     & 75.1 & 84.8 & 79.6 \\ \midrule
Distant Timex No Mask & 5k  & 57.9 & 67.0 & 62.1 \\ 
Distant Timex Masked  & 5k  & 61.9 & 71.5 & 66.4 \\ 
BeforeAfter No Mask   & 5k  & 51.5 & 59.5 & 55.2 \\ 
BeforeAfter Masked    & 5k  & 51.6 & 59.7 & 55.3 \\
\bottomrule
\end{tabular}
\caption{Transfer comparison for data with and without masking. Results are ensembled from three models trained in the same setting.}
\label{mask-effect}
\end{table}
%%%%%%%%%%%%%%%

%%%%%%%%%%%%%%%
\begin{table}[]
\small
\centering
\setlength{\tabcolsep}{3pt} % Default value: 6pt
\begin{tabular}{lr | lr | lr | lr}
\toprule
\multicolumn{2}{l}{\textbf{MATRES}} & \multicolumn{2}{l}{\textbf{DistantTimex}} & \multicolumn{2}{l}{\textbf{BeforeAfter}} & \multicolumn{2}{l}{\textbf{UDST 5k}}\\  \midrule

 said   & 16.0\%& won    & 2.02\%& was   &  2.60\% & is   & 5.8\%\\
 killed & 1.2\% & died   & 1.69\%& said  &  2.08\% & was  & 3.3\%\\
 found  & 1.0\% & said   & 1.67\%& came  &  1.70\% & have & 2.9\%\\
 says   & 0.9\% & began  & 1.41\%& is    &  1.06\% & are  & 2.5\%\\
 told   & 0.8\% & joined & 1.10\%& began &  0.99\% & be   & 2.2\%\\
 called & 0.8\% & took   & 1.10\%& made  &  0.85\% & get  & 1.5\%\\
 reported& 0.7\%& set    & 1.04\%& have  &  0.78\% & had  & 1.4\%\\
 saying & 0.7\% & killed & 1.04\%& left  &  0.77\% & know & 1.4\%\\
 say    & 0.7\% & born   & 0.96\%& had   &  0.77\% & do   & 1.1\%\\
 was    & 0.6\% & held   & 0.93\%& be    &  0.73\% & go   & 1.9\%\\ \bottomrule

\end{tabular}
\caption{Top 10 events for each dataset as a percentage of total event mentions.}
\label{tab:top-events}
\end{table}
%%%%%%%%%%%%%%%

\paragraph{Effect of Masking}
In Table \ref{mask-effect}, we test the effect of masking on model generalization. We train our model on the collected distant examples with and without masking, and report ensembled evaluation results on the MATRES test set. Our comparison shows that masking causes an increase in generalization for DistantTimex examples, but little change in BeforeAfter transfer, which still performs similar to the majority baseline.

\paragraph{Understanding the data distribution}
We report the most frequent events from each dataset in Table~\ref{tab:top-events}. MATRES is highly focused on reporting verbs, but the distant data has a much flatter distribution. BeforeAfter features more light verbs whereas DistantTimex features events with more complex semantics; possibly the model can learn more regular and meaningful patterns from such data, or relevant cues from a more similar event distribution (than found in BeforeAfter). We present event-label tuples in Appendix \ref{sec:most-common-tuples}.

\section{Related Work}

There is little direct prior work on using this kind of distant supervision for temporal relation extraction. Past work has studied automatic extraction of typical inter-event orderings \citep{chklovski-pantel-2004-verbocean,ning-etal-2018-improving, yao-huang-2018-temporal} to aid downstream temporal tasks, but these approaches represent events as single words (predicates) taken out of context, so the knowledge they can capture is limited. The commonsense acquisition method of \citet{zhou-etal-2020-temporal} learns more sophisticated information, but more about unary properties of events (typical time, duration) rather than relational knowledge. \citet{lin2020conditional} achieve a somewhat similar goal, but make a strong assumption about narrative-structured corpora and do not evaluate on in-context temporal relation extraction.

Our technique does not use a knowledge base like classic distant supervision methods \cite{Mintz:09}. However, because we eventually mask out the explicit temporal indicators, we are still using temporal information ``external'' to the final example to derive the label, hence why we invoke this term. A related concept is the idea of labeling functions \cite{ratner2016,hancock-etal-2018-training}, which are used to automatically construct training data for new domains. However, to our knowledge, these techniques have not been applied to temporal relation extraction, nor used in conjunction with masking as we do.

\section{Discussion}

We use explicit temporal cues to automatically identify examples of temporal relations between events in text. By masking these trivial features, a pre-trained Transformer model can learn from the remaining context and generalize to human-annotated benchmarks. Comparing performance for two distant labeling methods--using discourse connectives and linking events to time expressions--indicates that richer temporal cues exist in the second case. The scope of identified time expressions encompasses both explicit datetimes and natural language indicators (``now", ``recently", etc.).

While datetimes may be more common in news and historical articles, relative time expressions are present in diverse domains such as literature and colloquial texts. Where such indicators exist, our approach may be used to automatically collect distantly labeled temporal relations. More broadly, we believe that this label-and-mask paradigm could be used to collect targeted training data for a variety of NLP tasks.

\section*{Acknowledgments}

Thanks to the anonymous reviewers for their helpful comments. This material is also based on research that is in part supported by the Air Force Research Laboratory (AFRL), DARPA, for the KAIROS program under agreement number FA8750-19-2-1003. The U.S. Government is authorized to reproduce and distribute reprints for Governmental purposes notwithstanding any copyright notation thereon. The views and conclusions contained herein are those of the authors and should not be interpreted as necessarily representing the official policies or endorsements, either expressed or implied, of the Air Force Research Laboratory (AFRL), DARPA, or the U.S. Government.

\bibliography{eacl2021}
\bibliographystyle{acl_natbib}

\appendix
\section{Model Implementation}
\label{sec:impl}

As mentioned previously, we selected a set of hyperparameters by fine-tuning to approximate the performance of recent state of the art on MATRES, achieving an F1 of 79.8 with RoBERTa (base) and 80.3 with ELECTRA. Specifically, we arrived at a learning rate of 2e-5 with a warmup proportion of 0.1. Our batch size varied from 16-25 based on hardware.

Full implementation can be found at \url{https://github.com/xyz-zy/distant-temprel}

\section{Label Composition Across Datasets}
\label{sec:label-breakdown}

Table \ref{tab:label_percents} compares the label distribution of our distant data against the human-annotated MATRES dataset. In comparison, our datasets are lacking in \textsc{vague} relations (which have the lowest IAA in MATRES, and are particularly difficult to resolve) but emphasize the two most prominent classes. In few-shot settings, our model sees and trains on \textsc{vague} relations from MATRES.

%%%%%%%%%%%%%%%
\begin{table}[]
    \small
    \centering
    \begin{tabular}{l | c c c c }
    \toprule
     & \textsc{before} & \textsc{after} & \textsc{equal} & \textsc{vague}\\
     \midrule
    MATRES       & 34.5\% & 50.2\% & 3.5\% & 11.8\%  \\
    DistantTimex & 67.4\% & 31.4\% & 1.3\% & --      \\ 
    BeforeAfter  & 50.7\% & 49.3\% & --    & --      \\
     \bottomrule
     \end{tabular}
     \caption{Training data label distribution. Both of our distant datasets contain 10k examples; the MATRES training set contains 9.7k examples.}
     \label{tab:label_percents}
\end{table}
%%%%%%%%%%%%%%%

%%%%%%%%%%%%%%%
\begin{table}[]
\small
\centering
\setlength{\tabcolsep}{4pt} % Default value: 6pt
\begin{tabular}{ll l l l l  l  }
\toprule
  & \multirow{2}{*}{} &
\multicolumn{4}{c}{\textbf{MATRES}} & \textbf{$-$Vague} \\ 
\multicolumn{2}{l}{\textbf{Training Data}} & \textbf{Split} & \textbf{P} & \textbf{R} & \textbf{F1} & \textbf{Acc.}\\ \midrule
\multirow{2}{*}{Majority Label} & & Dev  & 52.6 & 60.2 & 56.1 & --\\
                              &      & Test & 50.7 & 58.6 & 54.3 & -- \\
                              \midrule
\multirow{2}{*}{MATRES}       &     & Dev  & 77.1 & 85.5 & 81.1  & 85.5 \\ 
                              &     & Test & 75.1 & 84.8 & 79.6  & 84.8 \\ \midrule
\multirow{6}{*}{DistantTimex} & 1k  & Dev  & 54.1 & 61.9 & 57.7 & 61.9 \\
                              & 1k  & Test & 50.8 & 58.7 & 54.5 & 58.7 \\
                              & 5k  & Dev  & 60.4 & 69.1 & 64.5 & 69.1 \\  
                              & 5k  & Test & 61.9 & 71.5 & 66.4 & 69.2 \\  
                              & 10k & Dev  & 64.0 & 73.2 & 68.3 & 73.2 \\
                              & 10k & Test & 61.5 & 71.1 & 66.0 & 71.1 \\ \midrule
\multirow{2}{*}{BeforeAfter}  & 10k & Dev  & 53.4 & 61.1 & 57.0 & 61.1 \\ %Epoch 3
                              & 10k & Test & 51.6 & 59.7 & 55.3 & 59.7 \\  \bottomrule
\end{tabular}
\caption{Expanded view of results on MATRES, comparing performance on only on \{\textsc{before}, \textsc{after}, \textsc{equals}\} examples (``$-$ Vague") versus the entire eval set. Presented results are majority-vote ensembled from three models trained in the same setting.}
\label{tab:tab-1-no-vague}
\end{table}
%%%%%%%%%%%%%%%

%%%%%%%%%%%%%%%
\begin{table}[]
\small
\centering
\setlength{\tabcolsep}{4pt} % Default value: 6pt
\begin{tabular}{ll l l l}
\toprule
  & \multirow{2}{*}{} &
            \multicolumn{2}{c}{\textbf{Full Test Set}} &\textbf{$-$Vague} \\ 
\textbf{Labeled Set} & \textbf{+Distant}               & \textbf{Acc.}           & \textbf{F1}           & \textbf{Acc.}    \\ \midrule
                            
\multirow{3}{*}{MATRES 1k}
                            & 1k   & 67.0 & 72.0 & 77.5  \\
                            
                            & 5k   & 69.2 & 75.0 & 79.1  \\ 
                            & 10k  & 70.1 & 75.5 & 80.8  \\ 
                            \midrule
\multirow{3}{*}{UDS-T 5k}
                            & 1k   & 58.1 & 62.3 & 67.1  \\
                            & 5k   & 65.9 & 70.7 & 76.2 \\
                            & 10k  & 67.9 & 71.8 & 78.5 \\
                          \bottomrule

\end{tabular}
\caption{Comparison of majority-vote ensembled performance on \{\textsc{before}, \textsc{after}, \textsc{equals}\} examples (``$-$Vague") versus performance on the entire test set. Performance is higher without vague examples, and increases with the number of DistantTimex examples added.}
\label{tab:no-vague}
\vspace{-2mm}
\end{table}
%%%%%%%%%%%%%%%

\section{Performance without \textsc{vague} Examples}

The \textsc{vague} label is defined to express indeterminacy and also has the lowest inter-annotator agreement in the MATRES dataset. These examples are also relatively more difficult for models to learn. Tables \ref{tab:tab-1-no-vague} and \ref{tab:no-vague} present an expanded view of our results, adding the evaluation accuracy on only \{\textsc{before}, \textsc{after}, \textsc{equals}\} examples. As expected, we observe that model performance increases across the board under this evaluation.

\section{Most Common Event-Label Tuples}
\label{sec:most-common-tuples}

Table \ref{tab:top-event-label-tuples} presents a comparison of the most common (event1, event2, label) tuples across datasets. In MATRES, the most common tuples are largely (event, ``said", \textsc{before}) events. The DistantTimex data features many examples of same-verb pairs (\emph{won-won}, \emph{died-died}). These examples frequently come from sentences or sentence pairs discussing related events of the same type, using dates to contrast them.

%%%%%%%%%%%%%%%
\begin{table}[]
\small
\centering
\setlength{\tabcolsep}{6pt} % Default value: 6pt
\begin{tabular}{l r }
\toprule
\multicolumn{2}{l}{\textbf{MATRES}} \\ \midrule 
 (`said', `said', \textsc{before})   & 1.69\% \\
 (`said', `said', \textsc{vague})    & 0.67\% \\
 (`said', `said', \textsc{after})    & 0.25\% \\
 (`said', `told', \textsc{before})   & 0.21\% \\
 (`called', `said', \textsc{before}) & 0.21\% \\
 (`killed', `said', \textsc{before}) & 0.16\% \\
 (`found', `said', \textsc{before})  & 0.15\% \\
 (`added', `said', \textsc{before}) & 0.15\% \\
 (`found', `said', \textsc{after}) & 0.14\% \\
 (`declined', `said', \textsc{before}) & 0.14\% \\
 \midrule
 
  \multicolumn{2}{l}{\textbf{DistantTimex}} \\ \midrule
  (`won', `won', \textsc{before})        & 0.44\%  \\
  (`died', `died', \textsc{before})      & 0.26\%  \\
  (`annexed', `seized', \textsc{before}) & 0.26\%  \\
  (`born', `graduated', \textsc{before}) & 0.25\%  \\
  (`crashes', `killed', \textsc{after})  & 0.20\%  \\
  (`killed', `torched', \textsc{after})  & 0.19\%  \\
  (`crashed', `crashed', \textsc{after}) & 0.17\%  \\
  (`died', `married', \textsc{before})   & 0.16\%  \\
  (`end', `remove', \textsc{before})     & 0.14\%  \\
  (`died', `killed', \textsc{before})    & 0.14\%  \\
 \bottomrule

\end{tabular}
\caption{Top 10 event-event-relation tuples per dataset as a percentage of total event mentions.}
\label{tab:top-event-label-tuples}
\end{table}
%%%%%%%%%%%%%%%

\section{Distant Examples}
\label{sec:distant-examples}

Table \ref{distant-exs} presents a sample of our distantly labeled data. Examples (a-b) show that our BeforeAfter parsing scheme can correctly identify linked events across sentence spans. Examples (c-d) display a variety in parsed syntactic structures that link events to timexes. 

%%%%%%%%%%%%%%%
\begin{table*}[t]
\centering
\small
\begin{tabular}{l p{11cm}ll}
\toprule
 & \textbf{Example Text} & \textbf{Heuristic} & \textbf{Label} \\
\midrule
(a) & A child prodigy, he \textbf{studied} music at the Julliard School of Music \underline{before} \textbf{turning} to art. & BeforeAfter  & \textsc{before}\\
\midrule
(b)&  Hill's support \textbf{came} only a few days \underline{after} former Prime Minister Paul Keating \textbf{made} a speech at the University of New South Wales in which he revived his campaign for a republic. & BeforeAfter  &  \textsc{after}\\
\midrule
(c) & \underline{1903}: General Electric \textbf{introduces} the first light set for the public sale for \$12, then the average weekly wage of a typical American worker. \underline{1962}: General Electric \textbf{designs} the National Christmas Tree for the first time. & DistantTimex & \textsc{before} \\ 
\midrule
(d) & Batista's EBX, the holding company for his five Rio-based companies, \textbf{increased} its workforce fivefold since \underline{2006} to 2,000 employees. State-run oil producer Petroleo Brasilero... plans to \textbf{add} 6,000 more by \underline{2013}. & DistantTimex & \textsc{before} \\ 
\midrule
(e) & It \textbf{took} until \underline{recently}, the official \textbf{said}, for everyone to realize ``it didn't work." & DistantTimex &  \textsc{before} \\ 
\bottomrule
\end{tabular}
\caption{Weakly labeled data examples, with the identifying heuristics underlined. Examples (a-b)  show different instances of labeling based on ``before/after" mentions. Examples (c-e) show instances of labeling event pairs based on anchoring time expressions, which range from datetimes to natural language indicators (e.g. ``recently"). While other timexes are identified in the examples shown, only the anchoring expressions for the induced labels are indicated.}
\label{distant-exs}
\end{table*}
%%%%%%%%%%%%%%%

\paragraph{Distant Example Masking}
\label{sec:masking}
Tables \ref{mask-beforeafter} and \ref{mask-distanttimex} present examples of our masking scheme on BeforeAfter and DistantTimex examples respectively.

Notably in Table \ref{mask-distanttimex}, multi-word timexes result in one mask token per word. All identified timexes in the examples are masked, even if they are not directly linked to the events in consideration.

%%%%%%%%%%%%%%%
\begin{table*}[p]
\centering
\small
\begin{tabular}{l p{6cm} p{6cm} l}
\toprule
 & \textbf{Example Text} & \textbf{Masked}  &  \textbf{Label} \\
\midrule
(a) & A child prodigy, he \textbf{studied} music at the Julliard School of Music \underline{before} \textbf{turning} to art. & A child prodigy, he \textbf{studied} music at the Julliard School of Music \underline{[mask]} \textbf{turning} to art. & \textsc{before}\\
\midrule
(b) &  Hill's support \textbf{came} only a few days \underline{after} former Prime Minister Paul Keating \textbf{made} a speech at the University of New South Wales in which he revived his campaign for a republic. &  Hill's support \textbf{came} only a few days \underline{[mask]} former Prime Minister Paul Keating \textbf{made} a speech at the University of New South Wales in which he revived his campaign for a republic.   &  \textsc{after}\\
\bottomrule
\end{tabular}
\caption{Masking of of weakly labeled data examples identified by before/after mentions. }
\label{mask-beforeafter}
\end{table*}
%%%%%%%%%%%%%%%

%%%%%%%%%%%%%%%
\begin{table*}[p]
\centering
\small
\begin{tabular}{l p{6cm} p{6cm} l}
\toprule
 & \textbf{Example Text} & \textbf{Masked} &  \textbf{Label} \\
\midrule

(a) & Batista's EBX, the holding company for his five Rio-based companies, \textbf{increased} its workforce fivefold since \underline{2006} to 2,000 employees. State-run oil producer Petroleo Brasilero has hired 22,000 employees in the past six years, ... and plans to \textbf{add} 6,000 more by \underline{2013}. & Batista's EBX, the holding company for his five Rio-based companies, \textbf{increased} its workforce fivefold since \underline{[mask]} to 2,000 employees. State-run oil producer Petroleo Brasilero has hired 22,000 employees in [mask] [mask] [mask] [mask], ... and plans to \textbf{add} 6,000 more by \underline{[mask]}.  & \textsc{before} \\
\midrule
(b) & It \textbf{took} until \underline{recently}, the official \textbf{said}, for everyone to realize ``it didn't work."& It \textbf{took} until \underline{[mask]}, the official \textbf{said}, for everyone to realize ``it didn't work."  &  \textsc{before} \\
\bottomrule
\end{tabular}
\caption{Masking of of weakly labeled data examples identified by anchoring timexes. }
\label{mask-distanttimex}
\end{table*}
%%%%%%%%%%%%%%%

\end{document}